\documentclass[runningheads]{llncs}

\usepackage{enumitem} % Used for enumerations and itemize
\usepackage{amsmath}
\usepackage{algpseudocode}

\usepackage[T1]{fontenc} % Used for displaying better typography in the tables.

\usepackage[labelfont=bf,font=small]{caption}
\usepackage[labelfont=bf,font=small]{subcaption}
\usepackage{graphicx}
\usepackage{hyperref}

\usepackage[separate-uncertainty=true]{siunitx}
\usepackage{booktabs,dcolumn,caption,tabularx,multirow,array}
\renewrobustcmd{\boldmath}{} % for avoiding siunitx using bold extended
\newrobustcmd{\B}{\fontseries{b}\selectfont} % abbreviation
\usepackage{amsmath}
\DeclareMathOperator*{\argmax}{arg\,max}

\usepackage[linesnumbered,vlined,ruled,commentsnumbered]{algorithm2e}
\SetKwRepeat{Do}{do}{while}

\SetCommentSty{mycommfont}

\begin{document}
% Title and Authors
\title{A Ring-Based Distributed Algorithm for Learning High-Dimensional Bayesian Networks}
\titlerunning{Ring-based distributed learning of high-dimensional BNs}
\author{Jorge D. Laborda\inst{1,2}\orcidID{0000-0002-6844-3970} \and
Pablo Torrijos\inst{1,2}\orcidID{0000-0002-8395-3848} \and
José M. Puerta\inst{1,2}\orcidID{0000-0002-9164-5191} \and
José A. Gámez\inst{1,2}\orcidID{0000-0003-1188-1117}
}
\authorrunning{JD. Laborda, P. Torrijos, JM Puerta and JA G\'amez}
\institute{Instituto de Investigaci\'on en Inform\'atica de Albacete (I3A). Universidad de Castilla-La Mancha. Albacete, 02071, Spain. \and
Departamento de Sistemas Inform\'aticos. Universidad de Castilla-La Mancha. Albacete, 02071, Spain.\\
\email{\{JorgeDaniel.Laborda,Pablo.Torrijos,Jose.Puerta,Jose.Gamez\}@uclm.es}}
\maketitle             

% Abstract and keywords
\begin{abstract}%   <- trailing '%' for backward compatibility of .sty file

Learning Bayesian Networks (BNs) from high-dimensional data is a complex and time-consuming task. Although there are approaches based on horizontal (instances) or vertical (variables) partitioning in the literature, none can guarantee the same theoretical properties as the Greedy Equivalence Search (GES) algorithm, except those based on the GES algorithm itself. In this paper, we propose a directed ring-based distributed method that uses GES as the local learning algorithm, ensuring the same theoretical properties as GES but requiring less CPU time. The method involves partitioning the set of possible edges and constraining each processor in the ring to work only with its received subset. The global learning process is an iterative algorithm that carries out several rounds until a convergence criterion is met. In each round, each processor receives a BN from its predecessor in the ring, fuses it with its own BN model, and uses the result as the starting solution for a local learning process constrained to its set of edges. Subsequently, it sends the model obtained to its successor in the ring. Experiments were carried out on three large domains (400-1000 variables), demonstrating our proposal's effectiveness compared to GES and its fast version (fGES).
\end{abstract}

\section{Introduction}
A Bayesian Network (BN) \cite{Jensen_Nielsen,Koller_Friedman,Pearl2014} is a graphical probabilistic model that expresses uncertainty in a problem domain through probability theory. BNs heavily rely on the graphical structure used to produce a symbolic (relevance) analysis \cite{relevance_BNs_1997}, which gives them an edge from an interpretability standpoint. The demand for explainable models and the rise of causal models make BNs a cutting-edge technology for knowledge-based problems.

A BN has two parts: A graphical structure that stores the relationships between the domain variables, such as (in)dependences between them, alongside a set of parameters or conditional probability tables that measure the weight of the relationships shown in the graph. Experts in the problem domain can help build both parts of the BN \cite{Kjaerulff_Madsen}. Unfortunately, this task becomes unsustainable when the scale of the problem grows. Nonetheless, learning BNs with data is a well-researched field, and even though learning the structure of a BN is an NP-hard problem \cite{learning-BN-NP_2004}, a variety of proposals have been developed to learn BNs from data \cite{chickering_optimal_2002,gamez_learning_2011,deCampos_2011,Scanagatta_review_2019}. Additionally, a number of studies have delved into high-dimensional problems \cite{tsamardinos_max-min_2006,structural_learning_arias_2015,Scanagatta_2015}.

The main focus of this paper is to address the problem of structural learning of BNs in high-dimensional domains to reduce its complexity and improve the overall result. To do so, we use a search and score approach within the equivalence class search space \cite{juanin:2013} while dividing the problem into more minor problems that can be solved simultaneously. Furthermore, our work exploits the advantages of modern hardware by applying parallelism to the majority of the phases of our algorithm.

To achieve these improvements, our research applies, as its core component, the recent proposal for BN fusion \cite{Puerta_2021} alongside an initial partitioning of all of the possible edges of the graph and the GES algorithm \cite{chickering_optimal_2002}. Therefore, in a few words, our algorithm starts by dividing the set of possible edges into different subsets and performing parallel learning of various networks, where each process is restricted to its according subset of edges. Once the batch has finished, the resulting BN is used as input for the following process, creating a circular system where the output of one process is the input of the following process. Our experiments were performed over the three largest BNs in the {\sf bnlearn} repository \cite{bnlearn}, showing that our algorithm reduces the time consumed while achieving good representations of these BNs.

The remainder of this paper is organized as follows: Section \ref{sec:preliminaries} provides a general introduction to BNs. Next, in Section \ref{sec:cGES}, our proposal is explained in detail. In Section \ref{sec:experiments}, we describe the methodology used to perform our experiments and present the results obtained. Finally, in Section \ref{sec:Conclusions}, we explain the conclusions we have arrived at throughout our work.

\section{Preliminaries}\label{sec:preliminaries}
\subsection{Bayesian Network}
A Bayesian Network (BN)\cite{Jensen_Nielsen,Koller_Friedman,Pearl2014} is a probabilistic graphical model frequently used to model a problem domain with predominant uncertainty. A BN is formally represented as a pair $\mathcal{B}=(G,\mathbf{P})$ where $G$ is a Directed Acyclic Graph (DAG) and $\mathbf{P}$ is a set of conditional probability distributions:

\begin{itemize}
    \item The DAG is a pair $G=(\mathbf{V}, \mathbf{E})$, where $\mathbf{V}=\{X_1,\dots X_n\}$ is the set of variables of the problem domain, and $\mathbf{E}$ is the set of directed edges between the variables: $\mathbf{E} = \{X \rightarrow Y \mid X \in \mathbf{V}, Y \in \mathbf{V}, X \neq Y\}$ $G$ is capable of representing the conditional (in)dependence relationships between $\mathbf{V}$ using the \textit{d-separation} criterion \cite{Pearl2014}.
    \item $\mathbf{P}$ is a set of conditional probability distributions that factorizes the joint probability distribution $P(\mathbf{V})$ by using the DAG structure $G$ and Markov's condition:

    \begin{equation}
        P(\mathbf{V}) = P(X_1,\dots,X_n) = \prod_{i=1}^{n}P(X_i|pa_G(X_i)),
    \end{equation}
    where $pa_G(X_i)$ is the set of parents of $X_i$ in $G$.
    
\end{itemize}

\subsection{Structural Learning of BNs} \label{subsec:structuralLearning}
Structural learning of BNs is the process of creating their DAG $G$ by using data\footnote{In this paper, we only consider the case of complete data, i.e., no missing values in the dataset}. This problem is an NP-hard problem \cite{learning-BN-NP_2004}; however, many solutions have been developed to learn BNs. We can classify these approaches into two groups: constraint-based and score+search solutions. In addition, some hybrid algorithms have also been developed (e.g., \cite{tsamardinos_max-min_2006,structural_learning_arias_2015}). The constraint-based algorithms use hypothesis tests to identify the conditional independences found in the data, while the score+search methods apply a search algorithm to find the best structure for a given score function or metric, which depends entirely on the given data. So, these approaches need a search method to find promising structural candidates and a scoring function to evaluate each candidate. We will only consider discrete variables and focus on the score+search methods.

We can see score+search methods as optimization problems where, given a complete dataset $D$ with $m$ instances over a set of $n$ discrete variables $\mathbf{V}$, the objective is to find the best DAG $G^*$ within the search space of the DAGs of the problem domain $G^n$, by means of a scoring function $f(G:D)$ that measures how well a DAG $G$ fits the given data $D$:

\begin{equation}
     G^*= \argmax_{G\in G^n} f(G:D)
\end{equation}

Different measurements have been used in the literature. The scoring functions can be divided into Bayesian and information theory-based measures (e.g., \cite{deCampos:scoring:2006}). Our work focuses on using the \emph{Bayesian Dirichlet equivalent uniform} (BDeu) score \cite{heckerman-1995}, but any other Bayesian score could be used in our proposal. This score is a particular case of BDe where a uniform distribution over all the Dirichlet hyperparameters is assumed.

\begin{equation}\label{eq:BDe}\small
BDeu(G\,|\,D) = log(P(G)) + \sum_{i=1}^{n} \left [ \sum_{j=1}^{q_i} \left[ log \left( \frac{\Gamma(\frac{\eta}{q_i})}{\Gamma(N_{ij} + \frac{\eta}{q_i})} \right) +  \sum_{k=1}^{r_i}  log \left( \frac{\Gamma(N_{ijk} + \frac{\eta}{r_iq_i})}{\Gamma(\frac{\eta}{r_iq_i})} \right) \right] \right], 
\end{equation}
where $r_i$ is the number of states for $X_i$, $q_i$ is the number of state configurations of $Pa_G(X_i)$, $N_{ij}$ and $N_{ijk}$ are the frequencies computed from data for maximum likelihood parameter estimation, $\eta$ is a parameter representing the equivalent sample size and $\Gamma()·$ is the \emph{Gamma} function.

A state-of-the-art algorithm for structural learning is the \emph{Greedy Equivalence Search} (GES) \cite{chickering_optimal_2002}. This algorithm performs a greedy approach over the equivalence space, using a scoring metric to search in two stages: \emph{Forward Equivalence Search} (FES) and \emph{Backward Equivalence Search} (BES). The FES stage is in charge of inserting edges into the graph, and when no further insertions improve the overall score, the BES stage begins to delete edges from the graph until there are no further improvements. It is proven that under certain conditions, GES will obtain an optimum BN representation of the problem domain. In our work, we use an alternative approach to GES, as described in \cite{juanin:2013}, where the FES stage is carried out in a totally greedy fashion while maintaining the BES stage intact. This improvement has been proven to be as effective as GES and to retain the same theoretical properties. To use this last algorithm as a control one, we implemented a parallel version of GES where the checking phase of the edges to add or delete is carried out in a distributed manner by using the available threads.

Apart from GES, we also consider  \emph{Fast Greedy Equivalence Search} (fGES) \cite{FGES:2017} as a competing hypothesis to test our proposal. fGES improves the original GES algorithm by adding parallel calculations. 

\subsection{Bayesian Network Fusion}
Bayesian Network Fusion is the process of combining two or more BNs that share the same problem domain. The primary purpose of the fusion is to generate a BN that generalizes all the BNs by representing all the conditional independences codified in all the input BNs. BN fusion is an NP-hard problem; therefore, heuristic algorithms are used to create an approximate solution \cite{pena_finding_2011}. To do so, the algorithm relies on a common ordering $\sigma$ of variables, and the final result depends strongly on the ordering $\sigma$ used.

In a recent work \cite{Puerta_2021}, a greedy heuristic method (GHO) is proposed to find a good ordering for the fusion process. To achieve a good order, GHO must find an order that minimizes the number of transformations needed. This is accomplished by using the cost of transforming a node into a sink throughout all DAGs, being used as a scoring method to evaluate orders and using it in a heuristic to find a good order.

\section{Ring-based distributed learning of BNs} \label{sec:cGES}
\begin{figure}[htbp]
    \centering
    \includegraphics[width=\textwidth]{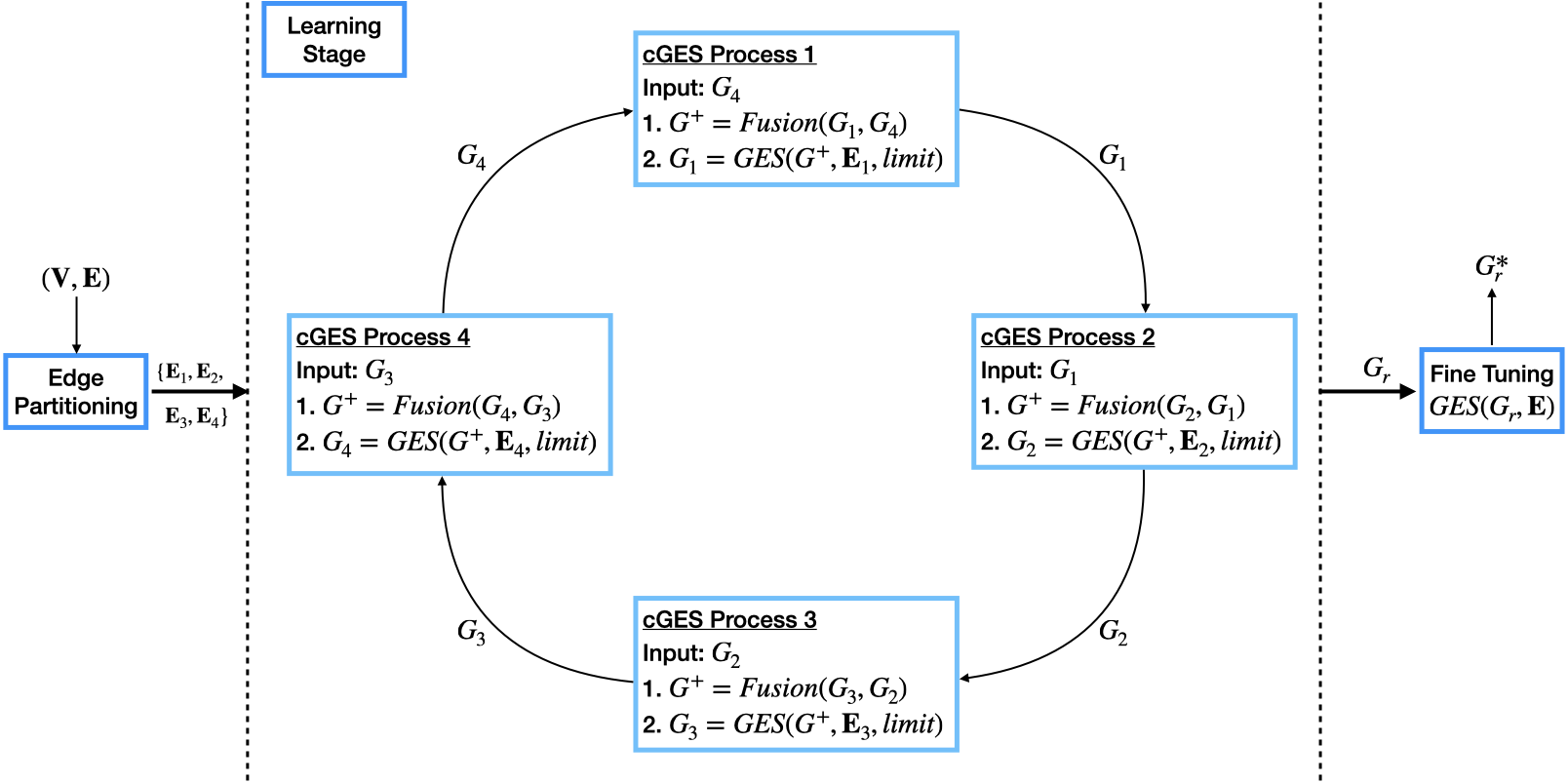}
    \caption{Graphical description of the proposed approach considering four processes}
    \label{fig:ring-fusion}
\end{figure}

Learning BNs for high-dimensional domains is a particularly complex process since it requires a much higher number of statistical calculations, which increases the iterations needed for the learning algorithms to converge. To reduce the computational demand of the learning process, we propose executing several simpler learning processes in parallel that reduce the time spent on the algorithm. We call our proposal \textit{Circular GES} (cGES); it is illustrated in Figure \ref{fig:ring-fusion}, and the scheme is depicted in Algorithm \ref{alg:cGES}.

\begin{algorithm}[h]
\setlength{\baselineskip}{1.3\baselineskip}

\caption{cGES($D$,$k$)}\label{alg:cGES}
\KwData{
$D$, dataset defined over $\mathbf{V} = \{X_1,\dots,X_n\}$ variables;

$k$, the number of parallel processes;

$l$, the limit of edges that can be added in a single GES process;
}

\KwResult{$G_r^*=(V,E)$, the resulting DAG learnt over the dataset $D$.}
\medskip

$\{\mathbf{E}_1, \dots, \mathbf{E}_k\} \leftarrow$ EdgePartitioning$(D,k)$

$go \leftarrow True$

$G_r \leftarrow \emptyset$

\For{($i=1,\dots, k$)}{
    $G_i \leftarrow \emptyset$
}

\While{$go$}{
    \tcc{Learning Stage}
    \For(in parallel){($i=1,\dots, k$)}{
    $\hat{G} \leftarrow Fusion.edgeUnion(G_i, G_{i-1})$
    
    $G_i \leftarrow GES(init = \hat{G}, edges = \mathbf{E}_i, limit= l, D)$
    }
    
    %$G^* \leftarrow BestDAG(\{G_1,\dots,G_k\})$   
    
    \tcc{Convergence Checking}
    $go \leftarrow False$
    
    \For{($i=1,\dots,k$)}{
        \If{($BDeu(G^*,D) - BDeu(G_i,D) \geq 0)$}{
            $G_r \leftarrow G_i$
            
            $go \leftarrow True$
        }
    }
}

\tcc{Fine Tuning}

$G_r^* \leftarrow GES(init=G_r,edges=\mathbf{E}, limit=\infty, D)$

\Return $G^*$

\end{algorithm}\smallskip

We can divide the algorithm into three stages:
\begin{enumerate}
    \item Edge partitioning. Given an input dataset $D$, with $\mathbf{V}=\{X_1,\dots,X_n\}$, as well as the set of possible edges $\mathbf{E} = \{X \rightarrow Y \mid X \in \mathbf{V}, Y \in \mathbf{V}, X \neq Y\}
    $, this step splits $\mathbf{E}$ into $k$ subsets $\mathbf{E}_1, \dots, \mathbf{E}_k$, such that $\mathbf{E} = E_1 \cup \dots \cup E_k$. This is done by using a \textit{score-guided complete-link hierarchical clustering} that partitions $\mathbf{E}$ into $k$ clusters of edges $\mathbf{E}_i$, where each possible edge can only be assigned to one and only one cluster of edges $\mathbf{E}_i$. First, we create $k$ clusters of variables by using the BDeu score (\ref{eq:BDe}) \cite{heckerman-1995} difference to measure the similarity or correlation between two variables: 
    
    \begin{equation}
        \label{eq:score-difference}
        s(X_i,X_j) = BDeu(X_i \leftarrow X_j \;| D) - BDeu( X_i \not\leftarrow X_j \;| D).
    \end{equation}
    
    Where, if $s(X_i, X_j)$ (\ref{eq:score-difference}) is positive, then adding $X_j$ as a parent of $X_i$ has an overall positive effect. The higher the score, the more related are the two variables. $s(X_i, X_j)$ is asymptotically equivalent to the mutual information. It's symmetric but non-negative, and it only measures the similarity of two variables, not the distance between them. We find a similar case in \cite{clustering_mutual_information_cite}. 
    
    To apply the complete link approach of the hierarchical clustering, we compute the similarity between clusters $C_r$ and $C_l$ as follows: 
    \begin{equation}
    s(C_r,C_l) = \frac{1}{|C_r|\cdot |C_l|} \sum_{X_i \in C_r} \sum_{X_j \in C_l} s(X_i,X_j)
    \end{equation}

    With the $k$ clusters of variables, we create the same number of clusters of edges. First, we assign all the possible edges among the variables of cluster $C_i$ to the subset $\mathbf{E}_i$. Next, we distribute all the remaining edges of variables belonging to different clusters. We attempt to balance the size of the resulting subsets by assigning the resulting edge with the end variables belonging to two clusters to the subset with the smallest number of edges. Finally, we obtain $k$ disjoint subsets of edges. The execution of this step occurs only once, at the beginning of the algorithm, and the resulting subsets are used to define the search space of each process of the learning stage.
    
    \item Learning stage. In this stage, $k$ processes learn the structure of a BN. Each process $i$ receives the BN learned by its predecessor ($i-1$) process and its $\mathbf{E}_i$ edge cluster as input. In every iteration, all the processes are executed in parallel, where each process is limited to their assigned $\mathbf{E}_i$ edge cluster. Each process works as follows: First, the process starts by carrying out a BN fusion \cite{Puerta_2021} between the predecessor's BN and the BN the process has learned so far. If it is the first iteration, the fusion step is skipped since no BNs have been learned yet, and we use an empty graph as starting point. Next, with the result of the fusion as a starting point, a GES algorithm is launched where the edges considered for addition and deletion are restrained to the edges of its $\mathbf{E}_i$ cluster. Furthermore, an additional option is to limit the number of edges that can be added in each iteration, resulting in a shorter number of iterations and avoiding introducing complex structures that would later be pruned during the merging process. After a preliminary examination, this limitation was set to  $(10 / k) \sqrt{n}$, ensuring that the limitation is tailored to the size of the problem, as well as to the number of subsets $\mathbf{E}_1, \dots, \mathbf{E}_k$ used.
    
    Once each process learns a BN, it is used as input for the next process, creating a ring topology structure. All the processes are independent and are executed in parallel. Each inner calculation needed by GES is also performed in parallel. As noted in the above section, we use the parallel version of GES, and all the processes store the scores computed in a concurrent safe data structure to avoid unnecessary calculations. Finally, when an iteration has finished, the convergence is checked by comparing whether any of the resulting BNs has improved its BDeu score over the best BN constructed so far. When no BN has outperformed the up to now best BN, the learning stage finishes.
    
    \item Fine tuning. Once the learning stage has finished, the parallel version of the GES algorithm is executed using the resulting BN as a starting point. This time, the GES algorithm uses all the edges of $\mathbf{E}$ without adding any limitation. As we expect to start from a solution close to the optimal, this stage will only carry out a few iterations. Since we apply a complete run of GES (FES+BES) over the resulting graph, all the theoretical properties of GES will be maintained as they are independent of the starting network considered.
\end{enumerate}

It is important to notice that, by using this ring topology, the fusion step only takes two networks as input, thus avoiding obtaining very complex (dense) structures and so reducing overfitting. Furthermore, throughout each iteration, the BNs generated by each process will be of greater quality, generalizing better with each iteration since more information is shared. By limiting the number of edges added, the complexity of each BN is smaller, and the fusions make smaller changes, creating more consistent BNs in each process. A general overview of the learning stage can be seen in Figure \ref{fig:ring-fusion}.

\section{Experimental evaluation}\label{sec:experiments}
This section describes the experimental evaluation of our proposal against competing hypotheses. The domains and algorithms considered, the experimental methodology, and the results obtained are described below.

\subsection{Algorithms} \label{subsec:algorithms}

In this study, we examined the following algorithms:
\begin{itemize}
    \item An enhanced version of the GES algorithm \cite{chickering_optimal_2002} introduced in \cite{juanin:2013} (see Section \ref{subsec:structuralLearning}). Notably, the implementation in this study incorporates parallelism to expedite the computational processes. In each iteration, to find the best edge to be added or deleted, the computation of the scores is implemented in parallel by distributing them among the available threads.

    \item The fGES algorithm, introduced in \cite{FGES:2017}.

    \item The proposed cGES algorithm (see Section \ref{sec:cGES}). We evaluate this algorithm with 2, 4, and 8 edge clusters, as well as limiting and non-limiting configurations for the number of edges inserted in each iteration.
\end{itemize}

\subsection{Methodology} \label{subsec:methodology}
Our methodology for evaluating Bayesian network learning algorithms involved the following steps:

First, we selected three real-world BNs from the Bayesian Network Repository in {\sf bnlearn} \cite{bnlearn} and sampled 11 datasets of \num{5000} instances for each BN. The largest BNs with discrete variables, namely \texttt{link}, \texttt{pigs}, and \texttt{munin}, were chosen for analysis. For each BN, Table \ref{tabla-BNs} provides information about the number of \textit{nodes}, \textit{edges}, \textit{parameters} in the conditional probability tables, the \textit{maximum} number of \textit{parents} per variable, average \textit{BDeu} value of the \textit{empty} network, and the structural Hamming distance between the \textit{empty} network and the moralized graph of the original BN (\textit{SMHD}) \cite{structural_distance:2009}.

\begin{table}[htb]
\caption{Bayesian networks used in the experiments.}
\label{tabla-BNs}
\resizebox{\textwidth}{!} {%
\begin{tabular*}{1.2\textwidth}{@{\extracolsep{\fill}}lS[table-format=4.0]S[table-format=4.0]S[table-format=7.0]S[table-format=1.0]S[table-format=4.4]S[table-format=4.0]}
\toprule
\multicolumn{1}{c}{\multirow{2}{*}{\textsc{\bfseries Network}}} &\multicolumn{6}{c}{\textsc{\bfseries Features}} \\
\cmidrule(){2-7}
& \multicolumn{1}{r}{\textsc{Nodes}} & \multicolumn{1}{r}{\textsc{Edges}} & \multicolumn{1}{r}{\textsc{Parameters}} & \multicolumn{1}{r}{\textsc{Max parents}} & \multicolumn{1}{r}{\textsc{Empty BDeu}} & \multicolumn{1}{r}{\textsc{Empty SMHD}}\\
\midrule
\textsc{Link}            & 724 & 1125 & 14211 & 3 & -410.4589 & 1739 \\
\textsc{Pigs}            & 441 & 592 & 5618 & 2 & -459.7571 & 806 \\
\textsc{Munin}           & 1041 & 1397 & 80592 & 3 & -345.3291 & 1843 \\
\bottomrule
\end{tabular*}
}
\end{table}

We considered several evaluation scores to assess the algorithms' efficiency and accuracy. These included the CPU time required by each algorithm for learning the BN model from data, the BDeu score \cite{heckerman-1995} measuring the goodness of fit of the learned BN with respect to the data normalized by the number of instances as in \cite{teyssier2005}, and the Structural Moral Hamming Distance (SMHD) between the learned and original BN, measuring the actual resemblance between the set of probabilistic independences of the moralized graph of the two models (see, e.g., \cite{Kim2019-moral}).

Our methodology tested the configuration of each algorithm on the 11 samples for each of the three BNs. The results reported are the average of these runs for each evaluation score. This approach allowed us to systematically evaluate the performance of the BN learning algorithms across multiple datasets and provide comprehensive insights into their efficiency and accuracy.

\subsection{Reproducibility} \label{subsec:reproducibility}  %%%%%%%%%%%%%%%%%%%%%%%
To ensure consistent conditions, we implemented all the algorithms from scratch, using Java (OpenJDK 8) and the Tetrad 7.1.2-2\footnote{\href{https://github.com/cmu-phil/tetrad/releases/tag/v7.1.2-2}{https://github.com/cmu-phil/tetrad/releases/tag/v7.1.2-2}} causal reasoning library. The experiments were conducted on machines equipped with Intel Xeon E5-2650 8-Core Processors and 64 GB of RAM per execution running the CentOS 7 operating system.

To facilitate reproducibility, we have made the datasets, code, and execution scripts available on GitHub\footnote{\href{https://github.com/JLaborda/cges}{https://github.com/JLaborda/cges}}. Specifically, we utilized the version 1.0 release for the experiments presented in this article. Additionally, we have provided a common repository on OpenML\footnote{\href{https://www.openml.org/search?type=data\&uploader\_id=\%3D\_33148\&tags.tag=bnlearn}{https://www.openml.org/search?type=data\&uploader\_id=\%3D\_33148\&tags.tag=bnlearn}} containing the 11 datasets sampled for each BN referencing their original papers.

\subsection{Results} \label{subsec:results}  %%%%%%%%%%%%%%%%%%%%%%%
Table \ref{tab:resultados} present the corresponding results for the BDeu score (\ref{tab:resultados}a), Structural Moral Hamming Distance (SMHD) (\ref{tab:resultados}b), and execution time (\ref{tab:resultados}c) of each algorithm configuration discussed in Section \ref{subsec:algorithms}. The notation \textsc{cGES-l} refers to the variant of \textsc{cGES} that imposes limitations on the number of added edges per iteration, while the numbers 2, 4, and 8 indicate the number of processes in the ring. The algorithm exhibiting the best performance for each Bayesian network is highlighted in bold to emphasize the superior results.

\begin{table}[ht]
\caption{Results (BDeu, SHMD and CPU Time)}
\label{tab:resultados}
\resizebox{\linewidth}{!} {%
\begin{tabular*}{1.4\textwidth}{@{\extracolsep{\fill}}lS[table-format=4.4]S[table-format=4.4]S[table-format=4.4]S[table-format=4.4]S[table-format=4.4]S[table-format=4.4]S[table-format=4.4]S[table-format=4.4]}
\toprule
\multicolumn{1}{c}{\multirow{2}{*}{\textsc{\bfseries Network}}} &\multicolumn{8}{c}{\textsc{\bfseries Algorithm}} \\
\cmidrule(){2-9}
& \multicolumn{1}{r}{\textsc{fGES}} & \multicolumn{1}{r}{\textsc{GES}} & \multicolumn{1}{r}{\textsc{cGES 2}} & \multicolumn{1}{r}{\textsc{cGES 4}} & \multicolumn{1}{r}{\textsc{cGES 8}} & \multicolumn{1}{r}{\textsc{cGES-l 2}} & \multicolumn{1}{r}{\textsc{cGES-l 4}} & \multicolumn{1}{r}{\textsc{cGES-l 8}}\\
\midrule
\textsc{Pigs}  & -345.1826 & \B -334.9989 & -335.6668 & -335.8876 & -335.5411 &      -335.1105 &      -335.1276 &      -335.1865 \\
\textsc{Link}  & -286.1877 & -228.3056 & -228.3288 & -227.1207 & \B -226.4319 &      -227.6806 &      -227.9895 &      -227.2155 \\
\textsc{Munin} & \B -186.6973 & -187.0736 & -187.1536 & -186.7651 & -187.8554 &      -186.9388 &      -187.2936 &      -187.4198 \\\midrule
\multicolumn{9}{c}{(a) BDeu score} \\
\bottomrule

\toprule
\multicolumn{1}{c}{\multirow{2}{*}{\textsc{\bfseries Network}}} &\multicolumn{8}{c}{\textsc{\bfseries Algorithm}} \\
\cmidrule(){2-9}
& \multicolumn{1}{r}{\textsc{fGES}} & \multicolumn{1}{r}{\textsc{GES}} & \multicolumn{1}{r}{\textsc{cGES 2}} & \multicolumn{1}{r}{\textsc{cGES 4}} & \multicolumn{1}{r}{\textsc{cGES 8}} & \multicolumn{1}{r}{\textsc{cGES-l 2}} & \multicolumn{1}{r}{\textsc{cGES-l 4}} & \multicolumn{1}{r}{\textsc{cGES-l 8}}\\
\midrule
\textsc{Pigs}  &  309.00 &  \B  0.00 &     31.00 &     36.91 &     21.00 &           4.36 &           4.18 &           5.18 \\
\textsc{Link}  & 1370.45 & 1032.36 &   1042.18 &    953.18 &    940.64 &      \B  937.91 &         952.64 &         941.55 \\
\textsc{Munin} & 1489.64 & \B 1468.45 &   1531.18 &   1521.38 &   1668.89 &        1503.25 &        1558.30 &        1623.22 \\\midrule
\multicolumn{9}{c}{(b) Structural Moral Hamming Distance (SHMD)} \\
\bottomrule

\toprule
\multicolumn{1}{c}{\multirow{2}{*}{\textsc{\bfseries Network}}} &\multicolumn{8}{c}{\textsc{\bfseries Algorithm}} \\
\cmidrule(){2-9}
& \multicolumn{1}{r}{\textsc{fGES}} & \multicolumn{1}{r}{\textsc{GES}} & \multicolumn{1}{r}{\textsc{cGES 2}} & \multicolumn{1}{r}{\textsc{cGES 4}} & \multicolumn{1}{r}{\textsc{cGES 8}} & \multicolumn{1}{r}{\textsc{cGES-l 2}} & \multicolumn{1}{r}{\textsc{cGES-l 4}} & \multicolumn{1}{r}{\textsc{cGES-l 8}}\\
\midrule
\textsc{Pigs}  &  \B  20.26 &  175.43 &    122.47 &    108.08 &    121.80 &          76.59 &          58.06 &          73.84 \\
\textsc{Link}  &  \B  41.12 &  746.54 &    694.08 &    463.92 &    447.62 &         383.04 &         276.72 &         286.56 \\
\textsc{Munin} & 12331.31 & 2000.00 &   1883.78 &   1330.62 &   1454.72 &        1433.19 &         895.76 &         \B 791.36 \\\midrule
\multicolumn{9}{c}{(c) CPU Time (seconds)} \\
\bottomrule
\end{tabular*}
}
\end{table}

These results lead us to the following conclusions:
\begin{itemize}
    \item Of the algorithms evaluated, \textsc{fGES} stands out as the least effective option, producing subpar results or exhibiting significantly longer execution times when obtaining a good result. In terms of the quality of the BN generated, \textsc{fGES} yields unsatisfactory outcomes, as evidenced by low BDeu scores and high SMHD values in the \texttt{pigs} and \texttt{link} networks.  Furthermore, when aiming to construct a reasonable network, \textsc{fGES} requires substantially longer execution times compared to both \textsc{GES} and all \textsc{cGES} variants. This is evident in the case of the \texttt{munin} network.

    \item Upon comparing the versions of cGES, namely \textsc{cGES-l} and \textsc{cGES}, which respectively impose limits on the number of edges that can be added by each FES run to $(10 / numClusters) \sqrt{nodes}$ and have no such restriction, it becomes evident that \textsc{cGES-l} outshines \textsc{cGES} in terms of performance. In most cases, \textsc{cGES-l} demonstrates superior performance in generating high-quality BNs compared to \textsc{cGES}. Additionally, it consistently achieves an impressive speed-up, with execution times reduced by approximately a half compared to \textsc{cGES}. These findings highlight the effectiveness of the edge limitation strategy employed in \textsc{cGES-l} and its significant impact on the learning process's quality and efficiency.

    \item When comparing the algorithms based on the number of ring processes (processes or edge subsets), it is challenging to establish a consistent pattern regarding the quality of the BNs generated. While there is a general trend of \textsc{cGES} performing slightly better with more partitions and \textsc{cGES-l} with fewer, this pattern may vary depending on the BN. However, regarding execution time, it is evident that using 4 or 8 clusters improves the efficiency compared to using 2 clusters. In particular, as the size of the BN increases, using 8 clusters tends to yield better execution times.

    \item Lastly, comparing the fastest variant of cGES in two out of three BNs (\textsc{cGES-l 4}) with \textsc{GES} yields noticeable speed improvements. \texttt{pigs}, \texttt{link}, and \texttt{munin} BNs experience speed-ups of 3.02, 2.70, and 2.23, respectively. These values are significant considering that both algorithms run parallel utilizing 8 CPU threads. Notably, the reduced speed-up execution time does not come at the cost of lower-quality BNs. In fact, \textsc{GES} performs better on \texttt{pigs} and \texttt{munin} BNs, while \textsc{cGES-l 4} excels with the \texttt{link} BN. However, these differences in performance are not as pronounced as those observed with the BN generated by \textsc{fGES} on the \texttt{pigs} and \texttt{link} networks.
\end{itemize}

\section{Conclusions}\label{sec:Conclusions}

Our study introduces cGES, an algorithm for structural learning of Bayesian Networks in high-dimensional domains. It employs a divide-and-conquer approach, parallelism, and fusion techniques to reduce complexity and improve learning efficiency. Our experimentation demonstrates that cGES generates high-quality BNs in significantly less time than traditional methods. While it may not always produce the absolute best solution, cGES strikes a favourable balance between BN quality and generation time. Another important point to be considered is that cGES exhibits the same theoretical properties as GES, as an unrestricted GES is run by taking the network identified by the ring-distributed learning process as its starting point.

As future works, the algorithm's modular structure opens up possibilities for applications such as federated learning \cite{federated_review}, ensuring privacy and precision.

\subsubsection{Acknowledgements} This work has been funded by the Government of Castilla-La Mancha and ``ERDF A way of making Europe'' under project SBPLY/21/180225/000062. It is also partially funded by MCIN/AEI/10.13039/501100011033 and ``ESF Investing your future'' through PID2019--106758GB--C33, TED2021-131291B-I00 and FPU21/01074 projects. Furthermore, this work has been supported by the University of Castilla-La Mancha and ``ERDF A Way of Making Europe'' under project 2023-GRIN-34437. Finally, this work has also been funded by the predoctoral contract with code 2019-PREDUCLM-10188, granted by the University of Castilla-La Mancha.

This preprint has not undergone peer review or any post-submission improvements or corrections. The Version of Record of this contribution is published in Lecture Notes in Computer Science, vol 14294, and is available online at \href{https://doi.org/10.1007/978-3-031-45608-4\_10}{https://doi.org/10.1007/978-3-031-45608-4\_10}.

\vskip 0.2in
\bibliographystyle{splncs04}
\bibliography{references}
\end{document}